\documentclass[letterpaper]{article} 
\usepackage{aaai2026}  
\usepackage{times}  
\usepackage{helvet}  
\usepackage{courier}  
\usepackage[hyphens]{url}  
\usepackage{graphicx} 
\usepackage{natbib}  
\usepackage{caption} 
\usepackage{algorithm}

\usepackage{newfloat}
\usepackage{listings}
\DeclareCaptionStyle{ruled}{labelfont=normalfont,labelsep=colon,strut=off} 
\floatstyle{ruled}
\newfloat{listing}{tb}{lst}{}
\floatname{listing}{Listing}
\usepackage{booktabs}
\usepackage{amsmath}
\usepackage{amssymb}
\usepackage{ragged2e}
\usepackage{lipsum} 
\usepackage{makecell}
\usepackage{multirow}

\newcommand{\method }{T$^2$Agent}

\usepackage{xcolor}
\usepackage{enumitem}

\definecolor{color1}{RGB}{250,128,114}
\definecolor{color2}{RGB}{237,140,59}  
\definecolor{color3}{RGB}{83,166,186}  
\definecolor{color4}{RGB}{72,126,186}

\usepackage{algpseudocode}
\usepackage[most]{tcolorbox}  
\usepackage{caption}       
\usepackage{listings}

\title{\method: A Tool-augmented Multimodal Misinformation Detection Agent with Monte Carlo Tree Search}

\author {
    Xing Cui\textsuperscript{\rm 1},
    Yueying Zou\textsuperscript{\rm 1},
    Zekun Li\textsuperscript{\rm 2},
    Peipei Li\textsuperscript{\rm 1}\thanks{Corresponding Author},
    Xinyuan Xu\textsuperscript{\rm 1},
    Xuannan Liu\textsuperscript{\rm 1},
    Huaibo Huang\textsuperscript{\rm 3}
}
\affiliations {
    \textsuperscript{\rm 1}Beijing University of Posts and Telecommunications\\
    \textsuperscript{\rm 2}University of California, Santa Barbara\\
    \textsuperscript{\rm 3}MAIS \& NLPR, Institute of Automation, Chinese Academy of Sciences\\
    \{cuixing, zouyueying2001, lipeipei, xuxinyuan, liuxuannan\}@bupt.edu.cn, \\zekunli@cs.ucsb.edu, huaibo.huang@cripac.ia.ac.cn
}

\usepackage{bibentry}

\begin{document}

\maketitle

\begin{abstract}
Real-world multimodal misinformation often arises from mixed forgery sources, requiring dynamic reasoning and adaptive verification. However, existing methods mainly rely on static pipelines and limited tool usage, limiting their ability to handle such complexity and diversity. 
To address this challenge, we propose \method, a novel misinformation detection agent that incorporates an extensible toolkit with Monte Carlo Tree Search (MCTS). The toolkit consists of modular tools such as web search, forgery detection, and consistency analysis. Each tool is described using standardized templates, enabling seamless integration and future expansion. To avoid inefficiency from using all tools simultaneously, a greedy search-based selector is proposed to identify a task-relevant subset. This subset then serves as the action space for MCTS to dynamically collect evidence and perform multi-source verification. To better align MCTS with the multi-source nature of misinformation detection, \method~ extends traditional MCTS with multi-source verification, which decomposes the task into coordinated subtasks targeting different forgery sources.  A dual reward mechanism containing a reasoning trajectory score and a confidence score is further proposed to encourage a balance between exploration across mixed forgery sources and exploitation for more reliable evidence. We conduct ablation studies to confirm the effectiveness of the tree search mechanism and tool usage. Extensive experiments further show that \method~ consistently outperforms existing baselines on challenging mixed-source multimodal misinformation benchmarks, demonstrating its strong potential as a training-free detector.
\end{abstract}
\begin{links}
    \link{Code}{https://github.com/cuixing100876/T2Agent}
\end{links}

\section{Introduction~\label{intro}}
\begin{figure*}
  \centering
    \includegraphics[width=\linewidth]{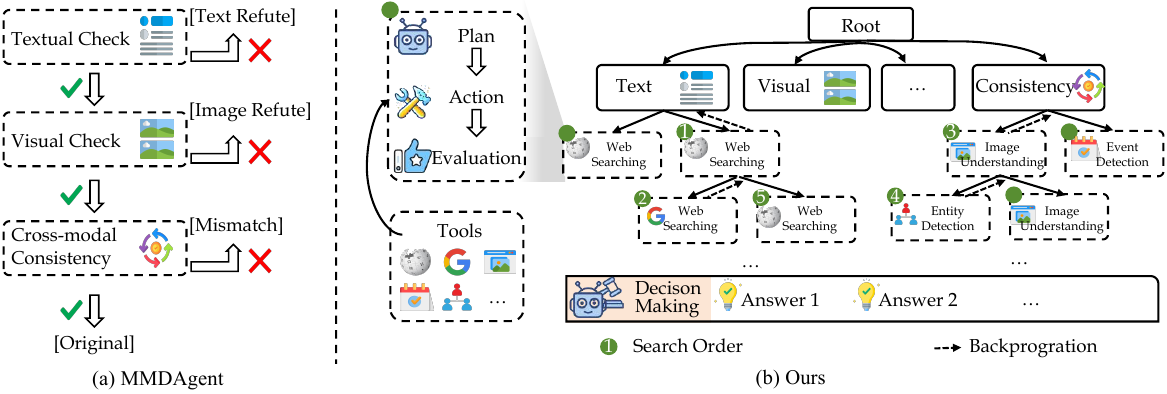}
    \caption{
    (1) MMDAgent adopts a fixed verification process. (2) Our \method~ builds a multi-source verification framework inspired by MCTS, enabling dynamic verification through adaptive tool selection and evidence integration.
    }
    \label{fig:intro-image}
\end{figure*}

The remarkable progress in Artificial Intelligence Generated Content (AIGC) technologies~\cite{ouyang2022training,touvron2023llama,dhariwal2021diffusion,rombach2022high,cui2024localize,cui2024instastyle} has lowered the barrier to generating sophisticated multimodal misinformation, posing severe threats to information integrity, public governance, economic stability, and societal well-being~\cite{zannettou2019web,apuke2021fake,allcott2017social}. 
Addressing this issue is not only a technical imperative but also a societal necessity. Therefore, developing approaches for multimodal misinformation detection is critical to safeguard the integrity of digital information ecosystems.

Automated multimodal misinformation detection is inherently complex, requiring capabilities in reasoning, information retrieval, and cross-modal verification. While recent automated approaches have made progress~\cite{liu2024mmfakebench,braun2024defame,beigi2024lrq}, they still fall short of emulating the dynamic strategies employed by human experts.
This gap can be attributed to two key factors. 
First, the diversity of forgery sources~\cite{guo2025each} necessitates tailored tools for different scenarios. For instance, the AMG benchmark~\cite{guo2025each} considers temporal consistencies, whereas MMfakebench~\cite{liu2024mmfakebench} incorporates counterfactual misinformation. 
However, most existing LLM-based methods rely on fixed and limited toolsets, which lack the flexibility and scalability required to handle such a wide range of forgery types.
Second, real-world multimodal misinformation often arises from mixed forgery sources, such as textual inaccuracies, image manipulations, or cross-modal inconsistencies~\cite{liu2024mmfakebench}.
Reliable detection requires not only the ability to perform in-depth exploitation to retrieve evidence for each forgery source but also to adaptively explore multiple potential forgery sources. However, existing LLM-based methods~\cite{liu2024mmfakebench,braun2024defame,beigi2024lrq,li2024large,lakara2024mad} typically rely on rigid designs that fail to strike this critical balance, limiting their effectiveness in complex scenarios.

To address these challenges, we introduce \method, a novel misinformation detection agent that incorporates an extensible toolkit with Monte Carlo Tree Search (MCTS). As illustrated in Fig.~\ref{fig:intro-image},  the extensible toolkit includes a range of functional tools such as web searching, time detection, forgery detection, counterfactual detection, image understanding, and entity recognition. Each tool is described using standardized templates, enabling seamless integration and future expansion for new tasks. Given that using all available tools simultaneously may overwhelm the agent and reduce efficiency~\cite{lu2025octotools}, we introduce a tool selection mechanism based on greedy search. This mechanism identifies the most relevant subset of tools for each task type, forming a task-specific subset. Then, the optimized subset serves as an action space for MCTS to dynamically collect evidence and perform multi-source verification.
Different from the previous MCTS, which is primarily designed for tasks with a single target, \method~ extends MCTS with multi-source verification. 
Specifically, \method~  first decomposes the misinformation detection task into multiple subtasks, each corresponding to a potential forgery source. Subsequently, it dynamically performs verification.
To guide this process effectively, we propose a dual evaluation function that combines two components: the reasoning trajectory score, which evaluates the quality of the reasoning path, and the result confidence score, which measures the certainty of the final decision.
By balancing exploration across mixed forgery sources with the exploitation of high-confidence evidence, \method~ enhances robustness and generalization in complex detection scenarios.

Our experiments demonstrate that \method~ presents a promising performance in mixed-source multimodal misinformation detection. On the MMfakebench~\cite{liu2024mmfakebench}, \method~ improves accuracy of the baseline MMDagent~\cite{liu2024mmfakebench} by 28.7\% with GPT4-o~\cite{chatgpt2022}, achieving a new state-of-the-art (SOTA). On AMG~\cite{guo2025each}, \method~ is competitive with existing training-based approaches, offering a promising direction for enhancing misinformation detection without the need of additional training. Ablation studies further confirm that the performance gains stem from the use of MCTS and tool integration. 
The primary contributions of this paper can be summarized as follows:
\begin{itemize}
\item We propose \method, a novel misinformation detection agent that integrates an extensible toolkit with Monte Carlo Tree Search (MCTS). By dynamically planning verification paths, leveraging tools for evidence collection, and performing structured multi-source validation, \method~ enables adaptive reasoning across diverse forgery sources.

\item We design an extensible toolkit based on modularized, standardized tool templates. This design facilitates rapid adaptation to various forgery patterns and allows seamless integration of new verification capabilities, improving system scalability and generalization.

\item We extend traditional MCTS with a multi-source verification framework by decomposing the detection task into subtasks, where each subtask targets a specific forgery source. A dual evaluation function further guides cross-subtask reasoning, balancing exploration across sources with exploitation of reliable evidence.

\end{itemize}

\section{Related work}
\subsection{Misinformation Detection}

Early misinformation detection datasets primarily focus on a single source, such as textual veracity distortion~\cite{wang2017liar,thorne2018fever,shu2020fakenewsnet,hanselowski2019richly,yao2023end,chen2023can},  or cross-modal inconsistency~\cite{luo2021newsclippings,aneja2023cosmos,shao2023detecting,suryavardan2023factify,nielsen2022mumin}. Some methods focus on mining forgery features from given content with specifically designed networks~\cite{hu2023learn,papadopoulos2025red,shao2023detecting}. 
Some recent works~\cite{cheung2023factllama,zeng-etal-2024-multimodal,liu2024fka,qi2024sniffer,yang2024take} fine-tune LLMs to enhance their detection capabilities for specific tasks. 
Benefiting from the powerful reasoning ability of LLMs, some works~\cite{kakizaki2025maft,li2024large,beigi2024lrq,lin2025fact,liu2025bidev} directly leverage LLM to build reasoning-driven misinformation detection pipelines. 
Some methods~\cite{khaliq-etal-2024-ragar,tahmasebi2024multimodal,braun2024defame,chen2024complex} further retrieve evidence to better handle the evolving nature of news content.

Despite being effective, these methods are unable to handle the situation where there are multiple sources of forged information in the real world. Several benchmarks~\cite{liu2024mmfakebench,guo2025each,wang2024mfc} are designed to focus on the scenarios where misinformation stems from combined sources. Although some works~\cite{lakara2024llm,dey2025fact} experiment on the multi-source forgery detection benchmark MMfakebench~\cite{liu2024mmfakebench}, they simply regarded the task as a binary classification of real or fake, ignoring the fine-grained identification for different sources of forged information.
To solve the challenge of mixed-source multimodal misinformation detection, MMDAgent~\cite{liu2024mmfakebench} integrates the reasoning ability of LLMs and relies on a predefined static detection workflow.
MGCA~\cite{guo2025each} extracts multi-view features and trains an end-to-end model to detect misinformation.
A recent work LRQ-FACT~\cite{beigi2024can}retrieves evidence by generating questions, and can detect multi-source fabricated information.
However, these methods are limited by the fixed workflows and a limited set of tools.
Our \method~, on the other hand, introduces a modified MCTS that adaptively searches across mixed forgery sources. This enables a principled trade-off between exploiting verified evidence and exploring diverse forgery sources, leading to improved performance in complex and heterogeneous misinformation settings.

\subsection{Monte Carlo Tree Search}
Monte Carlo Tree Search (MCTS)~\cite{coulom2006efficient} is a powerful heuristic search algorithm that has demonstrated exceptional performance in solving complex decision-making problems. Its core mechanism involves iteratively building a search tree and selecting the optimal action based on the outcomes of numerous simulations. The operation of MCTS typically consists of four key steps. \textbf{Selection}: Starting from the root node, a child node is recursively selected according to a policy such as the Upper Confidence bound applied to Trees (UCT) ~\cite{kocsis2006bandit} until a leaf node or a partially expanded node is reached. \textbf{Expansion}: If the selected node does not represent a terminal state, one or more new child nodes are created. \textbf{Simulation}: From the newly expanded node, a fast policy is executed to simulate a trajectory until a terminal state is reached. This simulation provides an estimate of the potential value of the node. \textbf{Backpropagation}: The outcome of the simulation is propagated back up the selection path to the root node, updating statistics such as visit counts and cumulative rewards for each node along the way. 

MCTS and its variants have demonstrated broad applicability across a wide range of domains, which have significantly improved decision-making performance in various areas~\cite{anthony2017thinking,ontanon2013combinatorial}.
Monte Carlo tree search simulates the human decision-making process. Some MCTS-based systems like AlphaGo~\cite{silver2016mastering} and AlphaZero~\cite{silver2017mastering} have even achieved superhuman performance. MCTS has been widely applied to domains such as robotic path planning~\cite{eiffert2020path} and combinatorial optimization~\cite{cazenave2009nested}. 
In recent years, researchers integrate MCTS with large language models to enhance the exploration of solution spaces in complex tasks such as question answering~\cite{yao2023tree,hao2023reasoning,xie2024monte,yu2023prompt,zhou2024language}, mathematics\cite{zhang2024accessing,chen2024alphamath}, websearch~\cite{yu2024improving,koh2024tree}, prompt optimization~\cite{wangpromptagent}, code generation~\cite{antoniades2024swe},  video understanding~\cite{yangdoraemongpt}, etc. 
These MCTS applications focus on tasks with a single target and are not well-suited for multi-source misinformation detection. In contrast,  we adapt MCTS to the mixed-source characteristics of multimodal misinformation detection by introducing a multi-source verification mechanism augmented with extensible tools.

\section{Method~\label{sec:method}}
\begin{figure*}
  \centering
    \includegraphics[width=\linewidth]{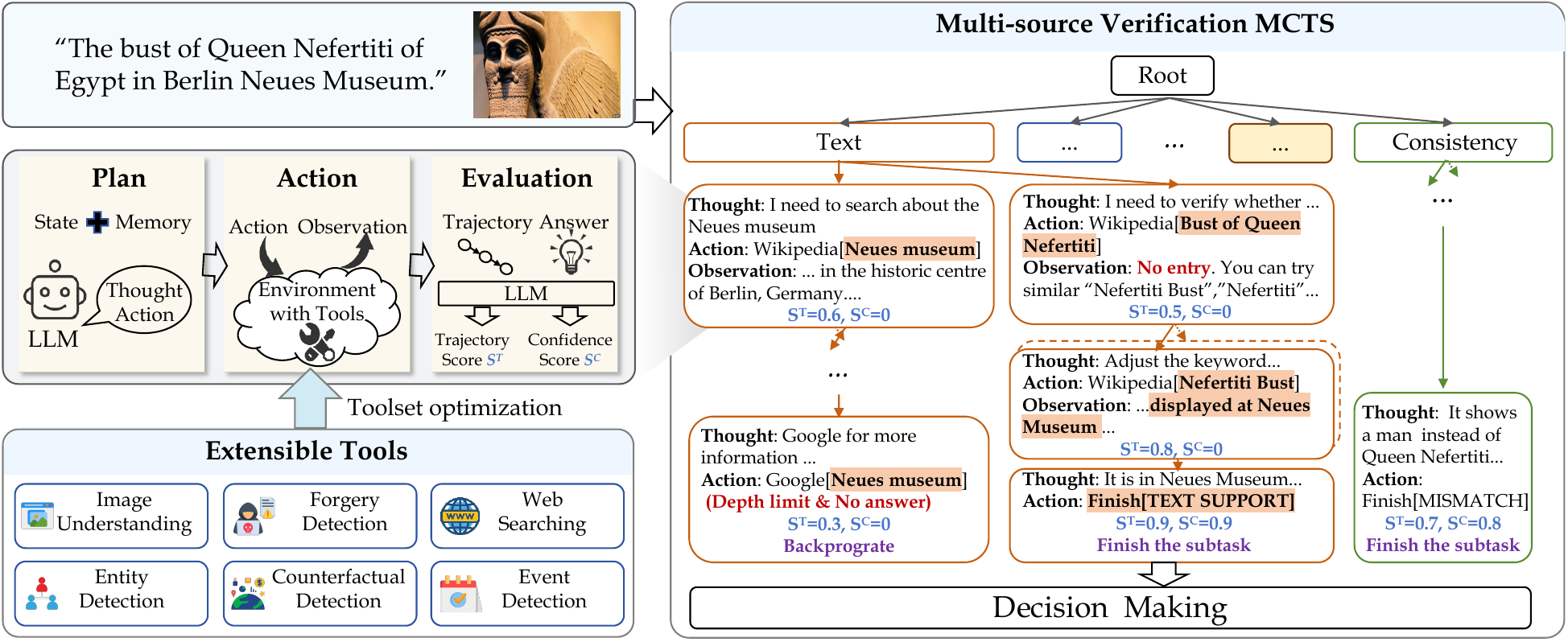}
    \caption{
Overview of \method. The toolkit acts as the action space, with a greedy search selecting relevant tools. \method~ extends MCTS via multi-source verification, breaking tasks into subtasks targeting different forgery sources. 
At each node of the tree search process, the agent plans verification paths, selects tools based on task requirements, and evaluates outcomes using a dual reward function that balances exploration across forgery sources with evidence exploitation.
    }
    \label{fig:main}
\end{figure*}

In this section, we first define the problem of multimodal misinformation detection (Sec.~\ref{sec:problem}). Then, we introduce our proposed framework, \method, which consists of two core components: a Multi-Source Verification Monte Carlo Tree Search (MCTS) and an extensible toolset for adaptive evidence reasoning.
Specifically, we extend traditional MCTS with multi-source verification (Sec.~\ref{sec:mcts}) to deal with potential mixed forgery sources in multimodal misinformation.
Our approach introduces a dual evaluation function and a custom backpropagation strategy that together maintain a balance between \texttt{exploration} and \texttt{exploitation} during search. Additionally, we propose a collaborative decision-making strategy to aggregate evidence and reach a final authenticity judgment.
Complementing the search process, our framework includes an extensible toolset (Section~\ref{sec:tool}) that supports plug-and-play integration of specialized tools. This enables dynamic and task-aware tool selection across diverse types of multimodal misinformation.

\subsection{Problem formulation\label{sec:problem}}
We address the task of mixed-source multimodal misinformation detection~\cite{liu2024mmfakebench}, which aims to assess the authenticity of online news content by analyzing multiple mixed sources of information and classifying it into various types of misinformation.
Unlike traditional single-source multimodal misinformation detection tasks that focus on generating a single correct answer, this task involves verifying content across different modalities and potential forgery sources.
It may require handling several sub-tasks, such as Textual Veracity Distortion, Visual Veracity Distortion, and Cross-modal Consistency Distortion~\cite{liu2024mmfakebench}.
Only through a comprehensive analysis of these multiple, mixed sources of evidence can the authenticity of a piece of news be reliably determined.

The verification process is a transition process among different states, where each state $s_t$ contains the thought, action $a_t$, and observation $o_t$. In our ~\method, the thought and action are determined by LLM. The observation is obtained by executing the action with the corresponding tool.
A misinformation detection agent begins by receiving an input instance $c$ that may contain text and visual content. Then, at each time step $t$, the agent generates an action $a_t \in \mathcal{A}$ based on the previous state $s_{t-1}$.
By executing $a_t$, the agent obtains a new observation $o_{t}$ and transitions to a new state $s_{t}$. This agent-environment interaction continues until either the agent makes a final judgment on the authenticity of the input information, or the maximum number of steps $T$ is reached.

\subsection{Multi-source Verification MCTS\label{sec:mcts}}
We propose Multi-source Verification MCTS, an extension of classical MCTS tailored for multi-source misinformation detection. During inference, similar to classic MCTS, it iteratively builds a search tree through the selection-expansion-simulation-backpropagation paradigm, predicting rewards at each step to guide the search.
Unlike traditional MCTS, we introduce subtask nodes in the search tree, each representing a potential forgery source (e.g., textual inconsistency, visual manipulation, or cross-modal discrepancy). The LVLM serves as the reasoning controller. At each node, it first makes a plan to generate a thought and corresponding action based on the current state and previous trajectory. We also utilized the trajectories of failures as memory, enabling drawing on past experiences. Once an action is selected, it invokes the corresponding tool from its extensible toolkit to retrieve external evidence or perform specific analyses. To balance exploration across mixed forgery sources with the exploitation of reliable evidence, we introduce a dual reward mechanism, which evaluates both the quality of the reasoning trajectory and the confidence of the final decision. This reward signal is used to update the value of its parent nodes during backpropagation, enabling dynamic adaptation to complex multimodal misinformation scenarios.

\textbf{Initialization}
The root of the tree represents the overall task of determining whether a given piece of news is true or false. The first layer of the tree consists of multiple child nodes, each representing a specific sub-task corresponding to mixed forgery sources, which can be adaptively pre-defined by the user based on the requirements of the task. 
To facilitate the exploration of the first step, we assign a weight to each child node.
We use LVLM to analyze the input content and estimate the probability distribution over these sub-task categories. This means that the model assigns a likelihood score to each sub-task, reflecting how relevant or necessary it is for verifying the claim. These probabilities guide the expansion of the tree by prioritizing which sub-tasks should be explored further.

\textbf{Evaluation}
To evaluate the nodes, we propose a dual scoring mechanism that combines a reasoning trajectory score with a result confidence score. Both scores are calculated by an LLM. We provide corresponding prompts in the appendix.
The reasoning trajectory score quantifies the overall quality and coherence of the reasoning process along the path from the root node to the current node. It is utilized in two scenarios. One is to evaluate the quality of the reasoning path during the simulation process and serve as the value of each step node, and the other is to calculate the quality of the reasoning path of the answer node and serve as part of the reward of the node.
The input to the reasoning trajectory score function includes all state-action pairs $\{s_i, a_i\}$ up to the current state.

{\footnotesize
\begin{equation}
S_t^T = LLM(\{s_i, a_i\}_{i=0 {\ldots} t}).
\end{equation}
}

Upon reaching a leaf node, we evaluate both the quality of the collected evidence and its internal consistency using the proposed confidence score. This score reflects the quality of the collected evidence and the degree to which the evidence supports the conclusion. The confidence score serves as the reward signal for the terminal node in the MCTS framework.
The input to the confidence scoring function includes all evidence, i.e., observations $\{o_i\}_{i=0 {\ldots} t}$ gathered along the current reasoning path, as well as the news $c$ to be verified. By analyzing the alignment and coherence among different pieces of evidence, it estimates a confidence value that captures the overall consistency of the verification process:

{\footnotesize
\begin{equation}
S_t^C = LLM(\{o_i\}_{i=0 {\ldots} t}, c).
\end{equation}
}

\textbf{Search Algorithm}
During the search process, our method automatically selects the sub-tasks that need verification. Specifically, we employ the Upper Confidence Bound for Trees (UCT) criterion~\cite{kocsis2006bandit}. At each iteration, the node with the highest UCT value is selected for expansion.
In this work, we revise the UCT formulation to better balance the trade-off between exploring under-visited sub-task nodes and exploiting those with high confidence scores. Our modification focuses on sub-task nodes that have not been explored, i.e., $N(s_i)=0$. Traditional methods usually assign an arbitrarily large bonus to unvisited nodes, which could lead to inefficient exploration. Different from them, we introduce a bias term and redefine the UCT function as follows:
{\footnotesize
\begin{equation}
UCT(s_t) = \frac{V(s_t)}{N(s_t)+1} + C\sqrt{\frac{\ln (N(s)+1)}{N(s_t)+1}}
\end{equation}
}
where $ V(s_t) = \alpha S_t^T + (1 - \alpha) S_t^C $ is the value of the terminal node. $ C $ is a hyperparameter. 
$ N(s_t) $ and $ N(s) $ are the visit count for the child node and the parent node, respectively.
By adding the bias term 1 to $ N(s_t) $, UCT can still be calculated when $N(s_t)=0$. The term $\sqrt{\frac{\ln (N(s)+1)}{N(s_t)+1}}$ allows the UCT values of unexplored nodes to be updated along with the overall situation. In the early stages of the search, in contrast to assigning an arbitrarily large bonus to unvisited nodes,  calculating UCT with our function avoids overly prioritized newly expanded sub-task nodes, thereby avoiding unnecessary resource allocation.
As the search progresses, the increasing visit count of the parent node gradually raises the UCT value of its unvisited children, thereby encouraging the exploration of unvisited sub-tasks.

Another key component of the search process is backpropagation. Considering the property that a piece of news can only be classified as true if all relevant modalities and sub-tasks independently confirm its authenticity,
we introduce a novel pruning strategy. If a sub-task node returns a high-confidence verification result indicating that the forgery source is likely true, we consider this sub-task completed and prune further expansion of its child nodes. By pruning already-confirmed branches, the search process avoids redundant exploration and allocates computational resources toward verifying remaining modalities. This strategy mimics how human experts operate: once a source or modality has been confirmed to be reliable, they tend to focus on validating other uncertain components.

\textbf{Decision Making}
After the MCTS search is completed, we perform a final decision-making step based on the verification results collected from all sub-task nodes.
If MCTS completes with a sub-task node with a high-confidence score indicating that the content is fake, the system classifies the entire news item as false, without further aggregation. This early-stop mechanism ensures efficient and decisive misinformation detection when strong evidence against authenticity is found in any modality.
In cases where no such high-confidence false signal is detected, we proceed to aggregate the results from all verified sub-task nodes using a heuristic probabilistic fusion strategy. For each sub-task (representing a potential forgery source), we calculate the probability that the model believes the input information belongs to each forgery source $i, (i=visual, text...)$ by utilizing the confidence score $S^{C,i}$. 
{\footnotesize
\begin{equation}
    p(\text{fake}^i) = 
    \begin{cases}
        S^{C,i}, & \text{if } \text{answer}(i) = \text{fake}, \\
        1 - S^{C,i}, & \text{if } \text{answer}(i) = \text{real}.
    \end{cases}
\end{equation}
}
We assume that different modalities (sub-tasks) provide independent evidence. Therefore, we define the overall likelihood that the news is true as:
{ \footnotesize
\begin{equation}
    p(real)=\prod_{i=1}^n{\left(1-  p\left(\text{fake}^i\right)\right)}^{\left(1/n\right)},
\end{equation}
}
where $n$ is the number of verified sub-tasks. We obtain the final result by comparing the probabilities:
{ \footnotesize
\begin{equation}
    answer=arg\max \left( p(real),\left\{ p(fake^i) \right\} _{i=1}^{n} \right) .
\end{equation}
}
The proposed multi-source verification MCTS enables the system to explore multiple sub-tasks in a structured manner, assess the quality of evidence collected along different paths, prune verified branches to focus on remaining uncertainties, and integrate each sub-task to obtain the final decision. It emulates the iterative and dynamic reasoning process of a human expert, which dynamically shifts between exploring new sources, validating evidence across modalities, and consolidating findings into a final decision.

\subsection{Extensible Toolset}\label{sec:tool}
Beyond the enhanced MCTS framework, \method~ incorporates a modular and extensible toolset designed to support diverse misinformation verification tasks across multiple forgery sources. This toolset serves as the action space within the tree search process, enabling the agent to invoke appropriate tools for evidence gathering.
Each tool is encapsulated in a tool card, which abstracts its functionality, input-output format, and invocation method into a unified structure. This modular representation allows for seamless integration and easy extension of new tools. The toolset includes web searching tools, time detection tools, forgery detection tools, counterfactual detection tools, image understanding tools, and entity recognition tools. Details are presented in the appendix.

While the full toolset provides comprehensive capabilities, not all tools are equally useful for every task~\cite{lu2025octotools}. To ensure both efficiency and effectiveness, we utilize an adaptive tool selection mechanism tailored to each benchmark.
We employ greedy search to efficiently search for the optimal subset of tools.
First, we predefine a minimal default toolset, denoted as $D_{\text{base}}$. Then, each candidate tool $d_i$ is evaluated by adding it to the base set.
We compute the accuracy improvement as $\Delta_{d_i} = \text{Acc}(D_{\text{base}} \cup \{d_i\}) - \text{Acc}(D_{\text{base}})$. If $\Delta_{d_i} > 0$, the tool $d_i$ is considered beneficial for the target task. The updated $D_{\text{base}}$ is:
\[
D_{\text{base}} = D_{\text{base}} \cup \{d_i\}.
\]
This adaptive configuration ensures that the system maintains high performance while minimizing computational overhead. 
It also enables \method~ to generalize across diverse domains.

\section{Experiments~\label{sec:exp}}

\begin{table}[htbp]
\centering
\caption{{Comparison results on MMfakebench.}}
\vspace{1mm}
\begin{tabular}{l|l|cccr}
\toprule
\textbf{Approach} & \textbf{Backbone} & \textbf{F1 $\uparrow$} & \textbf{Accuracy $\uparrow$} \\
\midrule
Standard Prompt & BLIP-2 & 0.167 & 0.328 \\
 & LLaVA-1.6 & 0.257 & 0.404 \\
 & GPT-4o & 0.492 & 0.609 \\
\midrule
MMD-agent & GPT-4.1-nano & 0.398 & 0.424 \\
 & GPT-4o-mini & 0.478 & 0.485 \\
 & GPT-4o & 0.614 & 0.616 \\
\midrule
LRQ-FACT & GPT-4o & 0.716 & 0.708\\
\midrule
Ours & GPT-4.1-nano & 0.568 & 0.569 \\
 & GPT-4o-mini & 0.631 & 0.629 \\
 & GPT-4o & \textbf{0.759} & \textbf{0.753} \\
\bottomrule
\end{tabular}

\label{tab:compare_mmfakebench}
\end{table}
\subsection{Experiment Setting}

\textbf{Datasets.} We evaluate \method~ on two challenging misinformation detection benchmark: MMFakeBench~\cite{liu2024mmfakebench} and AMG~\cite{guo2025each}. Both of them contain mixed-source misinformation and categorize news into multiple forgery classes. MMFakeBench~\cite{liu2024mmfakebench} contains 11,000 image-text pairs, covering ``Real", ``Textual Veracity Distortion (TVD)", ``Visual Veracity Distortion (VVD)", and ``Cross-modal Consistency Distortion (CCD)", with both human- and machine-generated images. We sample 1,000 validation instances, balanced as 300 Real, 300 TVD, 100 VVD, and 300 CMM. 
AMG~\cite{guo2025each} classifies news into five forgery categories: ``Image Fabrication", ``Non-evidential Image", ``Entity Inconsistency", ``Event Inconsistency", and ``Time Inconsistency".
The dataset contains 4,922 samples in total, divided into 3,532 for training, 517 for validation, and 973 for testing. The test set includes 575 Real samples, 74 Image Fabrication, 69 Non-evidential Image, 29 Entity Inconsistency, 136 Event Inconsistency, and 90 Time Inconsistency.

\textbf{Evaluation Metrics}.
To ensure a comprehensive comparison, we adopt the accuracy and the macro-F1 score as evaluation metrics. The macro-F1 score equally weighs performance across all classes by computing the harmonic mean of per-class precision and recall.

\subsection{Main Results}
\textbf{MMfakebench.} 
On MMfakebench, we compare with Standard Prompt, MMD-Agent~\cite{liu2024mmfakebench}, and LRQ-FACT~\cite{beigi2024can}. For Standard Prompt, we utilize BLIP-2~\cite{li2023blip}, LLaVA-1.6~\cite{liu2024llava}, and GPT-4o~\cite{chatgpt2022} as LLM backbones. Since the code of LRQ-FACT is not open-source, the results in the table are those from their paper. 
As shown in Table~\ref{tab:compare_mmfakebench}, our \method~ consistently outperforms all comparison methods. 
The results of the standard prompt indicate that the LLM itself has a certain ability to detect false information. This might be due to the fact that the LLM contains knowledge. However, this detection capability is insufficient.
Our \method~ outperforms the MMD-Agent under all three LLM models (GPT-4.1 Nano, GPT-4o Mini, and GPT-4o), achieving an average 28.7\% relative improvement in accuracy. 
Under a lightweight setting (GPT-4.1 Nano), \method~ boosts F1 by over 42.7\%, demonstrating strong generalization and robustness across model scales.
Notably, with GPT-4o Mini, our method achieves an accuracy of 0.629 and F1 of 0.631, which even outperforms MMD-agent with GPT-4o (accuracy 0.616, F1 0.614). \method~ also outperforms LRQ-FACT, demonstrating the necessity of dynamic verification.

A mismatch forgery case sampled from MM-FakeBench (Fig.~\ref{fig:main}) showcases the effectiveness of \method~ in performing multi-source verification through iterative exploration and adaptive pruning.
In the first iteration, the agent selects the text subtask node for investigation based on the initialization score. It fails to obtain conclusive evidence. We evaluate the node and update the value of itself and its parent nodes via backpropagation to reflect this uncertainty.
During the second iteration, the text node is reselected based on the evaluation score, and after tool invocation and simulation, the system confirms that the text is authentic, assigning it a high confidence score. Based on this reliable outcome, the corresponding branch is pruned to avoid redundant verification, demonstrating the method’s efficiency in identifying trustworthy sources.
In the third iteration, the agent shifts focus to a consistency forgery source node. Through structured reasoning and evidence aggregation, \method~ reaches a final judgment with high confidence and terminates the verification process successfully. This example highlights several key advantages of the framework: (1) the ability to iteratively refine node values based on partial evidence, preventing premature decisions; (2) the dynamic expansion of the search tree toward informative forgery sources; and (3) the integration of pruning strategies to improve computational efficiency without sacrificing accuracy.

\begin{table}[htbp]
\centering
\caption{Comparison with MMD-agent on AMG (grouped by backbone for easy comparison).}
\label{tab:compare_amg}
\begin{tabular}{l l c c}
\toprule
\textbf{Backbone} & \textbf{Approach} & \textbf{F1 $\uparrow$} & \textbf{Accuracy $\uparrow$} \\
\midrule
GPT-4.1-nano & MMD-agent & 0.290 & 0.192 \\
             & Ours      & \textbf{0.402} & \textbf{0.503} \\
\midrule
GPT-4o-mini  & MMD-agent & 0.360 & 0.227 \\
             & Ours      & \textbf{0.499} & \textbf{0.538} \\
\midrule
GPT-4o       & MMD-agent & 0.365 & 0.306 \\
             & Ours      & \textbf{0.510} & \textbf{0.579} \\
\bottomrule
\end{tabular}
\end{table}
\textbf{AMG.} 
We also conduct an evaluation of our method against the MMD-agent on the AMG dataset.
As shown in Table~\ref{tab:compare_amg}, our approach consistently outperforms the MMD-agent across all model sizes. In terms of F1 score, our method achieves improvements of 38.6\%, 38.6\%, and 39.7\% over the MMD-agent when using GPT-4.1-nano, GPT-4o-mini, and GPT-4o, respectively. 
The superior performance of our method can be attributed to its ability to handle the complexity of the AMG dataset, which contains five forgery sources. The MMD-Agent adopts a sequential decision-making strategy, which tends to suffer from error propagation. This means that incorrect judgments made at earlier stages affect subsequent decisions, leading to a degradation in overall performance.
In contrast, our approach explores a more balanced and holistic strategy by incorporating MCTS into the decision process. This allows our model to better assess the likelihood of each forgery type and avoid premature or biased decisions.

The results on MMfakebench and AMG collectively highlight the feasibility of constructing general-purpose, training-free misinformation detection agents using large language models and structured reasoning frameworks, paving the way for more adaptable verification systems.

\subsection{Ablation Study}
We conduct an ablation study on MM-FakeBench using the GPT-4.1-nano backbone over 1,000 samples to evaluate the effectiveness of the multi-source verification MCTS and toolset in our framework. Results are summarized in Table~\ref{tab:ablation}.

\textbf{MCTS search enables flexible evidence verification.}
MMD-Agent follows a rigid, rule-based verification path with fixed state transitions, limiting its adaptability when handling incomplete or low-confidence evidence across diverse forgery sources. In contrast, introducing MCTS-based tree search allows our agent to dynamically explore and prioritize subtasks based on real-time confidence estimates. This flexibility leads to notable improvements across all metrics: the F1 score increases from 0.398 to 0.535 (+34.4\%) and accuracy from 0.424 to 0.534 (+25.9\%).

\begin{table}[htbp]
\centering
\caption{Ablation Study on MMfakebench.}
\label{tab:ablation}
\begin{tabular}{l c c}
\toprule
\textbf{Method} & \textbf{F1 $\uparrow$} & \textbf{Accuracy $\uparrow$} \\
\midrule
MMD-agent (baseline)          & 0.398 & 0.424 \\
TOOLs  & 0.413	& 0.459 \\
MV\_MCTS           & 0.535 & 0.534 \\
MV\_MCTS+TOOLs (Ours) & \textbf{0.568} & \textbf{0.569} \\
\bottomrule
\end{tabular}
\end{table}

\textbf{Extensible tools enhance performance through adaptive reasoning.}
We conduct an ablation study to evaluate the effectiveness of our extensible toolset. The base configuration includes fundamental tools such as image understanding and Wikipedia lookup. Building upon this, we employ our tool selection mechanism to automatically identify and integrate the most beneficial additional tools—specifically, Google Search for up-to-date external knowledge and entity detection via the Baidu API\footnote{\url{https://ai.baidu.com/tech/imagerecognition/general}} for object categories in complex images. 
As shown in Table~\ref{tab:ablation}, using the selected tools alone yields gains over the baseline (F1: 0.413 vs. 0.398), confirming their utility. More importantly, when integrated into the MCTS framework, they lead to significant performance improvements: our full method achieves an F1 score of 0.568 and accuracy of 0.569, corresponding to relative gains of +6.2\% and +6.5\% over the MCTS-only variant. This indicates that our tool selection mechanism picks useful tools that integrate well with MCTS.

\subsection{Cost Analysis}
Table~\ref{tab:cost} compares the inference costs (in USD) between the MMD-Agent and our method on the MMFakeBench dataset, using three GPT-4 variants: GPT-4o, GPT-4o-mini and GPT-4.1-nano. 
While our method has higher computational costs with the same LLMs, it provides two key advantages:
(1) Strong adaptivity: Our \method~ features task-adaptive reasoning without the need of additional training. This capability is especially important in real-world mixed-source forgery detection. Besides, it allows the system to adapt to novel forgeries without requiring retraining when a new source of forgery emerges. 
(2) Better cost-effectiveness: As shown in Table~\ref{tab:compare_mmfakebench}, when using GPT-4o-mini, our  \method~ achieves an F1 score of 0.631, surpassing MMDAgent (F1 = 0.614) that relies on the larger GPT-4o model. Meanwhile, our method incurs significantly lower resource costs—only 129.4\$ compared to MMDAgent's 344.4\$. These results clearly indicate that our method offers both higher performance and greater cost-efficiency, showcasing a strong advantage in terms of cost-effectiveness.

\begin{table}[h]
\centering
\caption{{Cost comparison (USD) between MMD-agent and our \method~ on MMfakebench.}}
\begin{tabular}{l|cc}
\toprule
\textbf{Model} &\textbf{MMD-agent} &\textbf{Ours}\\
\midrule
GPT-4o &344.4 &1637.1 \\
GPT-4o-mini &14.3 &129.4 \\
GPT-4.1-nano &9.5 &76.2 \\
\bottomrule
\end{tabular}
\label{tab:cost}
\end{table}

\section{Conclusion}
In this paper, we introduce \method, a novel misinformation detection agent designed to tackle the complexities and diversities of mixed-source multimodal misinformation. To achieve this goal, we design an extensible toolkit incorporating various modular tools such as web search, forgery detection, and consistency analysis, and propose the integration of Monte Carlo Tree Search (MCTS) with multi-source verification capabilities.
The incorporation of these elements achieves a balance between exploration across mixed forgery sources and exploitation for more reliable evidence, enhancing the adaptability and efficiency of the detection process. 
Extensive results show the efficiency of our approach, suggesting a strong potential for our method as a training-free approach for enhancing detection accuracy in real-world applications.

\textbf{Limitations and future directions.} One notable limitation of our current implementation is the increased computational overhead introduced by the tree search mechanism. Further works may address this limitation by introducing efficient pruning strategies or designing hybrid approaches by combining the proposed method with lightweight expert models to guide the search more effectively. Besides, open-source toolchains introduce new security risks. In practice, these risks may be mitigated by implementing principles such as least privilege and whitelisting tool calls.

\noindent{\textbf{Acknowledgement}}
This research is sponsored by National Natural Science Foundation of China (Grant No. 62306041).

\bibliography{aaai2026}

@String(CVPR  = {CVPR})

@String(ECCV  = {ECCV})

@String(ICML  =	{ICML})

@String(ICLR  = {ICLR})

@String(IJCAI = {IJCAI})

@String(AAAI = {AAAI})

@String(EMNLP ={EMNLP})

@inproceedings{ouyang2022training,
  title={Training language models to follow instructions with human feedback},
  author={Ouyang, Long and Wu, Jeffrey and Jiang, Xu and Almeida, Diogo and Wainwright, Carroll and Mishkin, Pamela and Zhang, Chong and Agarwal, Sandhini and Slama, Katarina and Ray, Alex and others},
  booktitle={NeurIPS}, 
  year={2022}
}

@inproceedings{rombach2022high,
  title={High-resolution image synthesis with latent diffusion models},
  author={Rombach, Robin and Blattmann, Andreas and Lorenz, Dominik and Esser, Patrick and Ommer, Bj{\"o}rn},
  booktitle={CVPR},
  year={2022}
}

@inproceedings{dhariwal2021diffusion,
  title={Diffusion models beat gans on image synthesis},
  author={Dhariwal, Prafulla and Nichol, Alexander},
  booktitle={NeurIPS},
  year={2021}
}

@article{touvron2023llama,
  title={Llama: Open and efficient foundation language models},
  author={Touvron, Hugo and Lavril, Thibaut and Izacard, Gautier and Martinet, Xavier and Lachaux, Marie-Anne and Lacroix, Timoth{\'e}e and Rozi{\`e}re, Baptiste and Goyal, Naman and Hambro, Eric and Azhar, Faisal and others},
  journal={arXiv preprint arXiv:2302.13971},  
  year={2023}
}

@article{zannettou2019web,
  title={The web of false information: Rumors, fake news, hoaxes, clickbait, and various other shenanigans},
  author={Zannettou, Savvas and Sirivianos, Michael and Blackburn, Jeremy and Kourtellis, Nicolas},
  journal={JDIQ}, 
  year={2019}
}

@article{apuke2021fake,
  title={Fake news and COVID-19: modelling the predictors of fake news sharing among social media users},
  author={Apuke, Oberiri Destiny and Omar, Bahiyah},
  journal={TELEMAT INFORM}, 
  year={2021}
}

@article{allcott2017social,
  title={Social media and fake news in the 2016 election},
  author={Allcott, Hunt and Gentzkow, Matthew},
  journal={JEP},  
  year={2017}
}

@inproceedings{liu2024mmfakebench,
  title={Mmfakebench: A mixed-source multimodal misinformation detection benchmark for lvlms},
  author={Liu, Xuannan and Li, Zekun and Li, Peipei and Huang, Huaibo and Xia, Shuhan and Cui, Xing and Huang, Linzhi and Deng, Weihong and He, Zhaofeng},
  booktitle={ICLR},
  year={2024}
}

@inproceedings{braun2024defame,
  title={DEFAME: Dynamic Evidence-based FAct-checking with Multimodal Experts},
  author={Braun, Tobias and Rothermel, Mark and Rohrbach, Marcus and Rohrbach, Anna},
  booktitle={ICML},
year={2025}
}

@inproceedings{beigi2024lrq,
  title={Lrq-fact: Llm-generated relevant questions for multimodal fact-checking},
  author={Beigi, Alimohammad and Jiang, Bohan and Li, Dawei and Kumarage, Tharindu and Tan, Zhen and Shaeri, Pouya and Liu, Huan},
  booktitle={COLING},
  year={2025}
}

@inproceedings{guo2025each,
  title={Each Fake News is Fake in its Own Way: An Attribution Multi-Granularity Benchmark for Multimodal Fake News Detection},
  author={Guo, Hao and Ma, Zihan and Zeng, Zhi and Luo, Minnan and Zeng, Weixin and Tang, Jiuyang and Zhao, Xiang},
  booktitle={AAAI},
  year={2025}
}

@inproceedings{kocsis2006bandit,
  title={Bandit based monte-carlo planning},
  author={Kocsis, Levente and Szepesv{\'a}ri, Csaba},
  booktitle={ECML},
  year={2006}
}

@article{papadopoulos2025red,
  title={Red-dot: Multimodal fact-checking via relevant evidence detection},
  author={Papadopoulos, Stefanos-Iordanis and Koutlis, Christos and Papadopoulos, Symeon and Petrantonakis, Panagiotis C},
  journal={IEEE TCSS},
  year={2025}
}

@inproceedings{yao2023end,
  title={End-to-end multimodal fact-checking and explanation generation: A challenging dataset and models},
  author={Yao, Barry Menglong and Shah, Aditya and Sun, Lichao and Cho, Jin-Hee and Huang, Lifu},
  booktitle={SIGIR},
  year={2023}
}

@article{li2024large,
  title={Large language model agent for fake news detection},
  author={Li, Xinyi and Zhang, Yongfeng and Malthouse, Edward C},
  journal={arXiv preprint arXiv:2405.01593},
  year={2024}
}

@inproceedings{yang2024take,
  title={Take it easy: Label-adaptive self-rationalization for fact verification and explanation generation},
  author={Yang, Jing and Rocha, Anderson},
  booktitle={WIFS},
  year={2024}
}

@inproceedings{cheung2023factllama,
  title={Factllama: Optimizing instruction-following language models with external knowledge for automated fact-checking},
  author={Cheung, Tsun-Hin and Lam, Kin-Man},
  booktitle={APSIPA ASC},
  year={2023}
}

@inproceedings{zeng-etal-2024-multimodal,
    title = "Multimodal Misinformation Detection by Learning from Synthetic Data with Multimodal {LLM}s",
    author = "Zeng, Fengzhu  and
      Li, Wenqian  and
      Gao, Wei  and
      Pang, Yan",
    booktitle = "EMNLP",
    year = "2024"
}

@inproceedings{khaliq-etal-2024-ragar,
    title = "{RAGAR}, Your Falsehood Radar: {RAG}-Augmented Reasoning for Political Fact-Checking using Multimodal Large Language Models",
    author = "Khaliq, Mohammed Abdul  and
      Chang, Paul Yu-Chun  and
      Ma, Mingyang  and
      Pflugfelder, Bernhard  and
      Mileti{\'c}, Filip",
 
    booktitle = "FEVER",
    year = "2024"
}

@inproceedings{coulom2006efficient,
  title={Efficient selectivity and backup operators in Monte-Carlo tree search},
  author={Coulom, R{\'e}mi},
  booktitle={ICCG},
  year={2006}
}

@article{silver2017mastering,
  title={Mastering the game of go without human knowledge},
  author={Silver, David and Schrittwieser, Julian and Simonyan, Karen and Antonoglou, Ioannis and Huang, Aja and Guez, Arthur and Hubert, Thomas and Baker, Lucas and Lai, Matthew and Bolton, Adrian and others},
  journal={nature},
  year={2017}
}

@inproceedings{anthony2017thinking,
  title={Thinking fast and slow with deep learning and tree search},
  author={Anthony, Thomas and Tian, Zheng and Barber, David},
  booktitle={NeurIPS},
  year={2017}
}

@inproceedings{ontanon2013combinatorial,
  title={The combinatorial multi-armed bandit problem and its application to real-time strategy games},
  author={Ontan{\'o}n, Santiago},
  booktitle={AAAI},
  year={2013}
}

@article{silver2016mastering,
  title={Mastering the game of Go with deep neural networks and tree search},
  author={Silver, David and Huang, Aja and Maddison, Chris J and Guez, Arthur and Sifre, Laurent and Van Den Driessche, George and Schrittwieser, Julian and Antonoglou, Ioannis and Panneershelvam, Veda and Lanctot, Marc and others},
  journal={Nature},
  year={2016},
}

@inproceedings{yao2023tree,
  title={Tree of thoughts: Deliberate problem solving with large language models},
  author={Yao, Shunyu and Yu, Dian and Zhao, Jeffrey and Shafran, Izhak and Griffiths, Tom and Cao, Yuan and Narasimhan, Karthik},
  booktitle={NeurIPS},
  year={2023}
}

@article{zhang2024accessing,
  title={Accessing gpt-4 level mathematical olympiad solutions via monte carlo tree self-refine with llama-3 8b},
  author={Zhang, Di and Huang, Xiaoshui and Zhou, Dongzhan and Li, Yuqiang and Ouyang, Wanli},
  journal={arXiv preprint arXiv:2406.07394},
  year={2024}
}

@inproceedings{
chen2024alphamath,
title={AlphaMath Almost Zero: Process Supervision without Process},
author={Guoxin Chen and Minpeng Liao and Chengxi Li and Kai Fan},
booktitle={NeurIPS},
year={2024}
}

@inproceedings{
xie2024monte,
title={Monte Carlo Tree Search Boosts Reasoning via Iterative Preference Learning},
author={Yuxi Xie and Anirudh Goyal and Wenyue Zheng and Min-Yen Kan and Timothy P Lillicrap and Kenji Kawaguchi and Michael Shieh},
booktitle={NeurIPSW},  
year={2024},
}

@inproceedings{eiffert2020path,
  title={Path planning in dynamic environments using generative rnns and monte carlo tree search},
  author={Eiffert, Stuart and Kong, He and Pirmarzdashti, Navid and Sukkarieh, Salah},
  booktitle={ICRA},
  year={2020}
}

@inproceedings{cazenave2009nested,
  title={Nested Monte-Carlo search},
  author={Cazenave, Tristan},
  booktitle={IJCAI},
  year={2009}
}

@inproceedings{
lu2025octotools,
title={OctoTools: An Agentic Framework with Extensible Tools for Complex Reasoning},
author={Pan Lu and Bowen Chen and Sheng Liu and Rahul Thapa and Joseph Boen and James Zou},
booktitle={ICLRW},  
year={2025}
}

@inproceedings{wang2017liar,
  title={“Liar, Liar Pants on Fire”: A New Benchmark Dataset for Fake News Detection},
  author={Wang, William Yang},
  booktitle={ACL},
  year={2017}
}

@inproceedings{suryavardan2023factify,
  title={Factify 2: A multimodal fake news and satire news dataset},
  author={Suryavardan, S and Mishra, Shreyash and Patwa, Parth and Chakraborty, Megha and Rani, Anku and Reganti, Aishwarya Naresh and Chadha, Aman and Das, Amitava and Sheth, Amit P and Chinnakotla, Manoj and others},
  booktitle={AAAIW}, 
  year={2023}
}

@article{wang2024mfc,
  title={Mfc-bench: Benchmarking multimodal fact-checking with large vision-language models},
  author={Wang, Shengkang and Lin, Hongzhan and Luo, Ziyang and Ye, Zhen and Chen, Guang and Ma, Jing},
  journal={arXiv preprint arXiv:2406.11288},
  year={2024}
}

@article{shu2020fakenewsnet,
  title={Fakenewsnet: A data repository with news content, social context, and spatiotemporal information for studying fake news on social media},
  author={Shu, Kai and Mahudeswaran, Deepak and Wang, Suhang and Lee, Dongwon and Liu, Huan},
  journal={Big Data},
  year={2020},
}

@inproceedings{hanselowski2019richly,
  title={A richly annotated corpus for different tasks in automated fact-checking},
  author={Hanselowski, Andreas and Stab, Christian and Schulz, Claudia and Li, Zile and Gurevych, Iryna},
  booktitle={CoNLL},
  year={2019}
}

@inproceedings{chen2023can,
  title={Can llm-generated misinformation be detected?},
  author={Chen, Canyu and Shu, Kai},
  booktitle={ICLR},
  year={2024}
}

@inproceedings{luo2021newsclippings,
  title={Newsclippings: Automatic generation of out-of-context multimodal media},
  author={Luo, Grace and Darrell, Trevor and Rohrbach, Anna},
  booktitle={EMNLP},
  year={2021}
}

@inproceedings{aneja2023cosmos,
  title={Cosmos: Catching out-of-context misinformation with self-supervised learning},
  author={Aneja, Shivangi and Bregler, Chris and Nie{\ss}ner, Matthias},
  booktitle={AAAI},
  year={2023}
}

@inproceedings{shao2023detecting,
  title={Detecting and grounding multi-modal media manipulation},
  author={Shao, Rui and Wu, Tianxing and Liu, Ziwei},
  booktitle={CVPR},
  year={2023}
}

@inproceedings{hu2023learn,
  title={Learn over Past, Evolve for Future: Forecasting Temporal Trends for Fake News Detection},
  author={Hu, Beizhe and Sheng, Qiang and Cao, Juan and Zhu, Yongchun and Wang, Danding and Wang, Zhengjia and Jin, Zhiwei},
  booktitle={ACL},
  year={2023}
}

@inproceedings{liu2024fka,
  title={Fka-owl: Advancing multimodal fake news detection through knowledge-augmented lvlms},
  author={Liu, Xuannan and Li, Peipei and Huang, Huaibo and Li, Zekun and Cui, Xing and Liang, Jiahao and Qin, Lixiong and Deng, Weihong and He, Zhaofeng},
  booktitle={ACM MM},
  year={2024}
}

@inproceedings{kakizaki2025maft,
  title={MAFT: Multimodal Automated Fact-Checking via Textualization},
  author={Kakizaki, Kazuya and Matsunaga, Yuto and Furukawa, Ryo},
  booktitle={AAAI},
  year={2025}
}

@inproceedings{qi2024sniffer,
  title={Sniffer: Multimodal large language model for explainable out-of-context misinformation detection},
  author={Qi, Peng and Yan, Zehong and Hsu, Wynne and Lee, Mong Li},
  booktitle={CVPR},
  year={2024}
}

@article{lin2025fact,
  title={FACT-AUDIT: An Adaptive Multi-Agent Framework for Dynamic Fact-Checking Evaluation of Large Language Models},
  author={Lin, Hongzhan and Deng, Yang and Gu, Yuxuan and Zhang, Wenxuan and Ma, Jing and Ng, See-Kiong and Chua, Tat-Seng},
  journal={arXiv preprint arXiv:2502.17924},
  year={2025}
}

@inproceedings{liu2025bidev,
  title={Bidev: Bilateral defusing verification for complex claim fact-checking},
  author={Liu, Yuxuan and Sun, Hongda and Guo, Wenya and Xiao, Xinyan and Mao, Cunli and Yu, Zhengtao and Yan, Rui},
  booktitle={AAAI},
  year={2025}
}

@inproceedings{tahmasebi2024multimodal,
  title={Multimodal misinformation detection using large vision-language models},
  author={Tahmasebi, Sahar and M{\"u}ller-Budack, Eric and Ewerth, Ralph},
  booktitle={CIKM},
  year={2024}
}

@inproceedings{chen2024complex,
  title={Complex Claim Verification with Evidence Retrieved in the Wild},
  author={Chen, Jifan and Kim, Grace and Sriram, Aniruddh and Durrett, Greg and Choi, Eunsol},
  booktitle={ACL},
  year={2024}
}

@inproceedings{nielsen2022mumin,
  title={Mumin: A large-scale multilingual multimodal fact-checked misinformation social network dataset},
  author={Nielsen, Dan S and McConville, Ryan},
  booktitle={SIGIR},
  year={2022}
}

@inproceedings{hao2023reasoning,
  title={Reasoning with Language Model is Planning with World Model},
  author={Hao, Shibo and Gu, Yi and Ma, Haodi and Hong, Joshua and Wang, Zhen and Wang, Daisy and Hu, Zhiting},
  booktitle={EMNLP},
  year={2023}
}

@inproceedings{wangpromptagent,
  title={PromptAgent: Strategic Planning with Language Models Enables Expert-level Prompt Optimization},
  author={Wang, Xinyuan and Li, Chenxi and Wang, Zhen and Bai, Fan and Luo, Haotian and Zhang, Jiayou and Jojic, Nebojsa and Xing, Eric and Hu, Zhiting},
  booktitle={ICLR},
  year={2024}
}

@inproceedings{yu2024improving,
  title={Improving Autonomous AI Agents with Reflective Tree Search and Self-Learning},
  author={Yu, Xiao and Peng, Baolin and Vajipey, Vineeth and Cheng, Hao and Galley, Michel and Gao, Jianfeng and Yu, Zhou},
  booktitle={ICLR},
  year={2024}
}

@article{koh2024tree,
  title={Tree search for language model agents},
  author={Koh, Jing Yu and McAleer, Stephen and Fried, Daniel and Salakhutdinov, Ruslan},
  journal={arXiv preprint arXiv:2407.01476},
  year={2024}
}

@inproceedings{yu2023prompt,
  title={Prompt-Based Monte-Carlo Tree Search for Goal-oriented Dialogue Policy Planning},
  author={Yu, Xiao and Chen, Maximillian and Yu, Zhou},
  booktitle={EMNLP},
  year={2023}
}

@inproceedings{zhou2024language,
  title={Language agent tree search unifies reasoning, acting, and planning in language models},
  author={Zhou, Andy and Yan, Kai and Shlapentokh-Rothman, Michal and Wang, Haohan and Wang, Yu-Xiong},
  booktitle={ICML},
  year={2024}
}

@article{antoniades2024swe,
  title={SWE-Search: Enhancing Software Agents with Monte Carlo Tree Search and Iterative Refinement},
  author={Antoniades, Antonis and {\"O}rwall, Albert and Zhang, Kexun and Xie, Yuxi and Goyal, Anirudh and Wang, William},
  journal={arXiv preprint arXiv:2410.20285},
  year={2024}
}

@inproceedings{thorne2018fever,
title={FEVER: a large-scale dataset for fact extraction and VERification},
author={Thorne, James and Vlachos, Andreas and Christodoulopoulos, Christos and Mittal, Arpit},
booktitle={NAACL},
year={2018}
}

@inproceedings{yangdoraemongpt,
  title={DoraemonGPT: Toward Understanding Dynamic Scenes with Large Language Models (Exemplified as A Video Agent)},
  author={Yang, Zongxin and Chen, Guikun and Li, Xiaodi and Wang, Wenguan and Yang, Yi},
  booktitle={ICML},
  year={2024}
}

@article{lakara2024mad,
  title={MAD-Sherlock: Multi-Agent Debates for Out-of-Context Misinformation Detection},
  author={Lakara, Kumud and Sock, Juil and Rupprecht, Christian and Torr, Philip and Collomosse, John and de Witt, Christian Schroeder},
  journal={arXiv preprint arXiv:2410.20140},
  year={2024}
}

@article{chatgpt2022,
  title={Chatgpt},
  author={OpenAI},
  journal={https://openai.com/blog/chatgpt},
  year={2022}
}

@article{liu2024llava,
  title={Llava-next: Improved reasoning, ocr, and world knowledge},
  author={Liu, Haotian and Li, Chunyuan and Li, Yuheng and Li, Bo and Zhang, Yuanhan and Shen, Sheng and Lee, Yong Jae},
 journal= {https://llava-vl. github. io/blog/2024-01-30-llava-next},
  year={2024}
}

@article{beigi2024can,
  title={Can LLMs Improve Multimodal Fact-Checking by Asking Relevant Questions?},
  author={Beigi, Alimohammad and Jiang, Bohan and Li, Dawei and Tan, Zhen and Shaeri, Pouya and Kumarage, Tharindu and Bhattacharjee, Amrita and Liu, Huan},
  journal={arXiv preprint arXiv:2410.04616},
  year={2024}
}

@inproceedings{li2023blip,
  title={Blip-2: Bootstrapping language-image pre-training with frozen image encoders and large language models},
  author={Li, Junnan and Li, Dongxu and Savarese, Silvio and Hoi, Steven},
  booktitle={ICML},
  year={2023}
}

@article{lakara2024llm,
  title={LLM-Consensus: Multi-Agent Debate for Visual Misinformation Detection},
  author={Lakara, Kumud and Channing, Georgia and Sock, Juil and Rupprecht, Christian and Torr, Philip and Collomosse, John and de Witt, Christian Schroeder},
  journal={arXiv preprint arXiv:2410.20140},
  year={2024}
}

@article{dey2025fact,
  title={Fact-checking with contextual narratives: Leveraging retrieval-augmented llms for social media analysis},
  author={Dey, Arka Ujjal and Awan, Muhammad Junaid and Channing, Georgia and de Witt, Christian Schroeder and Collomosse, John},
  journal={arXiv preprint arXiv:2504.10166},
  year={2025}
}

@inproceedings{cui2024localize,
  title={Localize, understand, collaborate: Semantic-aware dragging via intention reasoner},
  author={Cui, Xing and Li, Peipei and Li, Zekun and Liu, Xuannan and Zou, Yueying and He, Zhaofeng},
  booktitle={NeurIPS},
  year={2024}
}

@inproceedings{cui2024instastyle,
  title={Instastyle: Inversion noise of a stylized image is secretly a style adviser},
  author={Cui, Xing and Li, Zekun and Li, Peipei and Huang, Huaibo and Liu, Xuannan and He, Zhaofeng},
  booktitle={ECCV},
  year={2024}
}

\appendix
\section{Appendix}\label{sec:Appendix} 
This appendix contains additional details for the submission entitled "\method: A Tool-augmented Multimodal Misinformation Detection Agent with Monte Carlo Tree Search".
The appendix is organized as follows:

\begin{itemize}
\item \S\ref{Appendix:toolset} Toolset Details.
\item \S\ref{Appendix:pseudocode} \method~ Pseudocode.
\item \S\ref{Appendix:experiment_details} Experiment Details.
\item \S\ref{Appendix:more_results} More Results.
\item \S\ref{Appendix:analysis} More Analysis.
\item \S\ref{Appendix:case_study} Case Study.
\item \S\ref{Appendix:prompt} Prompts.
\end{itemize}

\subsection{Toolset Details}\label{Appendix:toolset}
To effectively address complex verification tasks that require multi-modal misinformation detection, we design an extensible toolset for \method. We provide a detailed description of each tool in the toolset in this section.

\textbf{Web Searching}
This toolset is used to gather reliable knowledge from the web for information verification. We integrate the Google Search API~\footnote{https://developers.google.com/custom-search/v1/overview} and Wikipedia API~\footnote{https://www.wikipedia.org/}, which are particularly effective in retrieving general web content and encyclopedic knowledge, respectively.

\textbf{Time Detection}
This module is responsible for identifying the earliest appearance time of an image. To achieve this, we utilize TinEye~\footnote{https://tineye.com/}, a reverse image search engine, to trace the earliest online publication date of a given visual content.

\textbf{Forgery Detection}
This toolset includes pre-trained forgery detection models designed to detect digital manipulations. The system supports the integration of new forgery detection models to meet the user's needs. For the AMG~\cite{guo2025each} benchmark, we employ PSCC-NET(Liu et al. 2022), a powerful manipulation detection network, to identify subtle image tampering.

\textbf{Counterfactual Detection}
Unlike forgery detection, which focuses on identifying whether media has been digitally altered or synthesized (e.g., deepfakes), this module detects inconsistencies between the depicted scene and real-world facts, i.e., counterfactual content. We use LLava-34B~\cite{liu2024llava}, a powerful vision-language model, to analyze images for logically or physically implausible elements.

\textbf{Image Understanding}
This component provides a detailed, question-driven interpretation of visual content. In our implementation, the image understanding model is aligned with the backbone model of the overall framework.

\textbf{Entity Recognition}
This tool enables precise identification of key entities within visual content, such as people, landmarks, and organizations. We use the Baidu Entity Recognition API~\footnote{https://ai.baidu.com/tech/imagerecognition/general}, which specializes in extracting entities from images.

\subsection{\method~ Pseudocode}\label{Appendix:pseudocode}
Our method \method~ consists of a Multi-source Verification MCTS and a Tool Selection mechanism.
Alg.~\ref{alg:mcts} presents the pseudocode of Multi-source Verification MCTS. Our method begins by initializing a root node representing the task of verifying news authenticity, with child nodes corresponding to various forgery sources weighted by an LVLM analysis. 
During each iteration, promising nodes are selected using a modified UCT criterion to balance exploration and exploitation effectively. 
Upon selecting a specific node, the agent initiates a simulation process. During this simulation, the agent iteratively performs reasoning steps (thoughts), executes actions, and collects observations. This iterative reasoning continues until either a conclusive answer is found or the predefined maximum search depth 
$d$ is reached.
At the termination of the simulation, an evaluation phase assesses the entire simulation process using a dual scoring mechanism that integrates both reasoning trajectory scores and confidence scores. The results from this evaluation are then backpropagated to update the values of the relevant nodes.
Finally, the decision-making component aggregates the verification results from the updated nodes to produce a final determination on the authenticity of the news.

\begin{algorithm*}[ht]
\centering
\caption{Multi-source Verification MCTS \label{alg:mcts}}
\begin{minipage}{0.9\linewidth} 
\small
\begin{algorithmic}
    \Require Initial state $s$, LVLM model $p_{LVLM}$, number of generated actions $n$, depth limit $d$, number of simulations $K$, exploration weight $C$, hyperparameter $\alpha$
    \State Initialize the tree with sub-task nodes. \Comment{Initialization}
    \For {$k \gets 1, \dots, K$} \Comment{Simulation Loop}
        \State $s \gets$ root node
        
        \State Select node$
            s \gets \arg\max_{s' \in children(s)} \left[V(s') + C\sqrt{\frac{\ln (N(s)+1)}{N(s')+1}}\right]
            $\Comment{Selection}

        \While{$s$ not terminal and depth $< d$} \Comment{Expansion and simulation}
            \If{$s$ has unvisited children}
                \State Expand $s$ by generating $n$ new child nodes using $p_{LVLM}$
            \EndIf
            \State Select next node $s' \in children(s)$ using modified UCT: 
        \EndWhile

        \State Combine scores: $V(s) = \alpha S_t^T + (1-\alpha) S_t^C$ \Comment{Evaluation}
        
        \For {$t \gets T - 1, \dots, 0$} \Comment{Backpropagation}
            \State {$V(s_t)\leftarrow \frac{V(s_t)(N(s_t) ) + V(s)}{N(s_t)+1}$}
        \EndFor
    \EndFor
    
    \If{Early stop with an answer} \Comment{Decision making}
        \State return answer
    \Else
        \State     $p(real)=\prod_{i=1}^n{ \left(1- p\left(\text{fake}^i\right)\right)}^{\left(1/n\right)}$
    \State $answer=arg\max \left( p(real),\left\{ p(fake^i) \right\} _{i=1}^{n} \right) $
    \EndIf
\end{algorithmic}
\end{minipage}
\end{algorithm*}

Although the full set of tools offers broad functionality, not all tools are equally valuable for every task~\cite{lu2025octotools}. To balance efficiency and performance, we leverage greedy search to systematically identify the most effective subset of tools. Alg.~\ref{alg:mcts} presents the pseudocode of the tool selection mechanism. In practice, we select a subset (100 samples) of all data samples for evaluation function $f$ to save time.
It is worth to notice that our method is differnet from OctoTools~\cite{lu2025octotools}.
OctoTools evaluates each tool independently by adding it to a fixed base toolset and measuring its individual contribution. In contrast, our method incrementally adds tools one by one in a greedy fashion, evaluating the combined effect at each step. 
We conduct experiments to compare our method with OctoTools~\cite{lu2025octotools}. Our approach employs Web Searching, Counterfactual Detection, and Entity Recognition, while Octo-Tools additionally incorporates Image Understanding. Experimental results show that our method achieves an F1 score of 0.568, outperforming Octo-Tools (F1 = 0.550). This suggests that although Image Understanding can be beneficial when used in isolation, its integration into a multi-tool framework may lead to performance degradation. One possible explanation is that Image Understanding can partially substitute for other tools—such as Entity Recognition—but with lower accuracy. This redundancy and reduced precision may interfere with the verification process, ultimately harming performance.
The results shows that our strategy enables a more accurate assessment of tool selection.

\begin{algorithm*}[ht]
\centering
\caption{Tool selection \label{alg:bo}}
\begin{minipage}{0.9\linewidth} 
\small
\begin{algorithmic}
    \State \textbf{Input:} Toolbox $\mathcal{D} = \{d_i\}_{i=1}^n$, base toolset $\mathcal{D}_\text{base}$
    \State \textbf{Output:} Optimized toolset $\mathcal{D}^*$
    
    \State $\text{Acc}_\text{baseline} \gets \text{Acc}(\mathcal{D}_\text{base})$  \Comment{Stage 1: Baseline setup}

    \For {each $d_i$ in $D$ \textbf{such that} $d_i \notin \mathcal{D}_\text{base}$} \Comment{Stage 2: Individual tool evaluation}
        \State $\mathcal{D}_i \gets \mathcal{D}_\text{base} \cup \{ d_i \}$
        \State $\text{Acc}_i \gets \text{Acc}(\mathcal{D}_i)$
        \State $\Delta_{d_i} \gets \text{Acc}_i - \text{Acc}_\text{baseline}$
        \If{$\Delta_{d_i} > 0$}
            \State $\mathcal{D}_\text{base} \gets \mathcal{D}_i$
        \EndIf
    \EndFor
    
    \State $\mathcal{D}^* \gets \mathcal{D}_\text{base}$ \Comment{Stage 3: Select optimized toolset}
    
    \State \textbf{Return} $\mathcal{D}^*$

\end{algorithmic}
\end{minipage}
\end{algorithm*}

\subsection{Experiment Details}\label{Appendix:experiment_details}
 In our implementation, we set the number of sampled nodes to $n = 2$ and the exploration weight to $C = 2$. Additionally, the hyperparameter $\alpha=0.5$ is used to balance the reasoning trajectory score and the result confidence score in the evaluation.

\subsection{More Results}\label{Appendix:more_results}

We present the results on the AMG benchmark in Table~\ref{tab:compare_amg_mmdagent}. It is important to note that the AMG dataset is initially designed for supervised learning. All compared methods, including CAFE (Chen et al. 2022), MCAN (Wu et al. 2021), and MGCA~\cite{guo2025each}, are trained or fine-tuned on the provided training set. In contrast, our proposed \method~ is based on tool-augmented tree search with large language models, requiring no training or fine-tuning.
Specifically, \method~ with GPT-4o obtains an F1 score of 0.510, surpassing CAFE (Chen et al. 2022) and MCAN (Wu et al. 2021). Notably, the test set of AMG has a highly imbalanced class distribution. Under such conditions, the F1 score is a particularly meaningful metric, and our method's competitive performance on the F1 score suggests its power without any task-specific training.
Despite underperforming compared to MGCA~\cite{guo2025each}, which achieves the highest F1 score of 0.567, our method demonstrates superiority in the following aspects: (1) Rapid adaptation and low-cost deployment: 
Although training-based methods can achieve strong performance, they (Chen et al. 2022; Wu et al. 2021; Guo et al. 2025) typically require designing different model architectures for different scenarios and often suffer from poor out-of-domain generalization. In contrast, our \method~ is training-free and can handle emerging misinformation without any additional training, making it particularly valuable in the early stages of information outbreaks.
(2) Data privacy and regulatory compliance:
In domains with strict privacy and compliance requirements—such as healthcare or finance—fine-tuning on task-specific training data may be infeasible or even prohibited. Our method circumvents this issue by avoiding task-specific training altogether.
(3) Benefit from powerful LLMs:
As shown in the results, as the model size increases (from GPT-4.1-nano to GPT-4o), the performance of \method~ consistently improves, rising from  0.503 to 0.579. This indicates that the performance may be further improved with a more powerful model.

\begin{table*}[h]
\centering
\caption{{Comparision results on AMG.}}
\vspace{-2mm}
\begin{tabular}{l|ccc|ccc}
\toprule
   & \multicolumn{3}{c|}{\textbf{Training based}}                                  & \multicolumn{3}{c}{\textbf{Training free}}   \\ \cline{2-7} 
   & \multirow{2}{*}{\textbf{CAFE}} & \multirow{2}{*}{\textbf{MCAN}} & \multirow{2}{*}{\textbf{MGCA}} & \multicolumn{3}{c}{\textbf{Ours}}            \\ \cline{5-7} 
   &                       &                      &                       & GPT-4.1-nano & GPT-4o-mini & GPT-4o \\ 
\midrule
\textbf{F1 $\uparrow$}  & 0.467                 & 0.461                & 0.567                 &0.402         &0.499           & 0.510  \\
\textbf{Accuracy $\uparrow$}  & 0.638                 & 0.612                & 0.739                 & 0.503        &0.538            & 0.579  \\

\bottomrule
\end{tabular}
\label{tab:compare_amg_mmdagent}
\end{table*}

\begin{table*}[h]
\centering
\caption{{Comparison with different settings of MMD-agent on MMfakebench.}}
\vspace{-2mm}
\begin{tabular}{l|ccc|c}
\toprule
               & \multicolumn{3}{c|}{\textbf{MMD-agent}}                                           & \multirow{2}{*}{\textbf{Ours}} \\ \cline{2-4}
               & Match-Text-Image & Image-Match-Text & {Text-Image-Match (default)} &                       \\ \midrule
\textbf{F1 $\uparrow$}       & 0.448            & 0.480             & 0.614                              & \textbf{0.759}                \\
\textbf{Accuracy $\uparrow$}       & 0.497            & 0.503            & 0.616                              & \textbf{0.753}               \\
\bottomrule
\end{tabular}
\label{tab:mmd_setting}
\end{table*}

We also evaluate the MMD-agent with different verification order settings to make a fair comparison. Experiments are conducted on the MMFakeBench dataset with GPT-4o as the Large Vision-Language Model (LVLM). We set three different verification order settings of the MMD-Agent: Match-Text-Image, Image-Match-Text, and Text-Image-Match (the default setting). As shown in Table~\ref{tab:mmd_setting}, the performance of the MMD-Agent varies considerably across these settings, revealing a strong dependence on the order in which forgery sources are processed.
In contrast, our method dynamically explores multiple, interleaved forgery sources without being constrained or biased by any predefined processing order. As a result, our approach achieves consistently superior performance across all metrics, demonstrating its robustness and effectiveness.

\begin{table}[h]
\centering
\caption{Comparison of different initialization methods (Experiment with GPT-4.1-nano).}
\label{tab:mmd_initialization}
\begin{tabular}{l c c c}
\toprule
\textbf{Method} & \textbf{F1 $\uparrow$} & \textbf{Accuracy $\uparrow$} & \textbf{Iterations $\downarrow$} \\
\midrule
Random          & 0.560 & 0.560 & 2.40 \\
Ours            & \textbf{0.568} & \textbf{0.569} & \textbf{2.28} \\
\bottomrule
\end{tabular}
\end{table}

\subsection{More Analysis}\label{Appendix:analysis}
\textbf{Initialization analysis}
During the initialization phase, we use a LVLM to assess and score different nodes. These scores determine the priority order for exploring sub-nodes, i.e., nodes with higher scores will be selected and explored first.
To investigate the impact of different initialization strategies, we compare our LVLM-based initialization method against a random initialization approach. As shown in Table~\ref{tab:mmd_initialization}, the overall performance metrics—including F1 score and accuracy are very similar between the two methods. For example, both achieve nearly identical F1 scores (0.560 vs. 0.568), and the differences in accuracy (0.560 vs. 0.569) are minimal. These results suggest that the choice of initialization strategy has little effect on final performance, indicating that our method is robust and capable of fully exploring all possible forgery sources regardless of which node is explored first.
However, a key advantage of our LVLM-based initialization is its ability to begin exploration from more promising forgery sources. Consequently, our approach requires fewer average iterations (2.28) to reach a conclusion compared to random initialization (2.40). This reduction in the number of exploration steps demonstrates that starting from a promising node enhances efficiency.

\subsection{Case Study}\label{Appendix:case_study}
\textbf{Good case} The good case demonstrates that our method avoids early termination and generates a correct answer only after gathering reliable evidence. At the 0-th iteration, the system correctly predicts "mismatch" but, due to low confidence, continues to verify. In the 1st iteration, it explores alternative forgery sources, and in the 2nd iteration, it further confirms the "mismatch" hypothesis. Once sufficient evidence is collected, the system returns the final result.

\textbf{Bad case} Despite employing MCTS and an extensible toolset to achieve more effective exploration and exploitation, our method can still be limited by the capabilities of individual tools. In this bad case, the model incorrectly deemed mismatch cross-modal information as consistent, resulting in a wrong final judgment. Similar issues may arise from weaknesses in forgery detection, counterfactual reasoning, web searching, etc. Future work may alleviate this problem by using more powerful tools.

\begin{figure*}[ht]
\begin{tcolorbox}[
    colback=white, 
    colframe=green!60!black, 
    fonttitle=\bfseries,
    title=Good case: Output a correct answer after obtaining reliable evidence,
    enhanced,
    attach boxed title to top left={yshift=-2mm,xshift=4mm},
    boxed title style={
            colback=green!60!black,
        colframe=green!60!black, 
    },
    coltitle=white 
]
\noindent
\begin{tabular}{@{}c@{}}
    \includegraphics[width=0.3\textwidth]{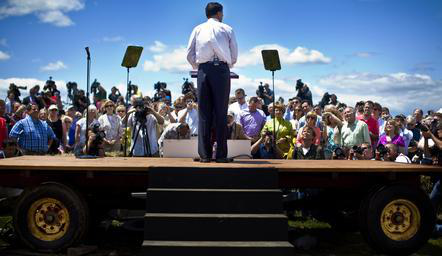}  
\end{tabular}
\hfill
\begin{tabular}{p{0.45\textwidth}}
    \textbf{News Text}: Romney announces Ryan as his running mate in front of the USS Wisconsin on Aug 11, 2012, in Norfolk, Va.\\
    \textbf{Binary label}: {Fake}\\
    \textbf{Multiple label}: {Mismatch}\\
\end{tabular}
\begin{lstlisting}[breaklines=true，basicstyle=\ttfamily\footnotesize, columns=fullflexible]
=======================Iter 0: select node <match>=====================
Thought 1: I need to determine what the image depicts, focusing on whether it shows the USS Wisconsin a political event with Romney and Ryan.
Action 1: VQA[What is shown in the image?]
Observation 1: The image shows a person standing on a stage or platform facing a large crowd of people. The crowd appears to be attending a gathering or event outdoors with some individuals taking photos or recording. The sky above is partly cloudy with patches of blue. The stage is elevated with steps leading up to it and there are two microphones or speakers positioned on either side of the person on stage.
Thought 2: Given the description the image depicts a person on stage at an outdoor event with a crowd, but there is no clear indication of the USS Wisconsin or specific political figures like Romney or Ryan. The scene appears to be a public gathering possibly a rally or speech but without identifying features linking it to the event described in the text.
Action 2: Finish[MISMATCH]
reward = 0.2
=======================Iter 1: select node <text>======================
Thought 1: The news mentions Romney announcing Ryan as his running mate on August 11, 2012 in Norfolk, Virginia, in front of the USS. To verify this, I should check credible information regarding Romney's 2012 vice-presidential pick and the specific event location and date.
Action 1: Google[Romney Ryan 2012 Norfolk USS Wisconsin]
Observation 1: Retrieved Information 1: Romney Announces Ryan as VP Running Mate Aug 11 2012 ... ... WI as his vice-presidential running mate during a campaign event at the retired battleship USS Wisconsin in Norfolk Virginia August 11 2012.     
    Retrieved Information 2: Mitt Romney campaigns with Paul Ryan Aug 13 2012 ...     
    Retrieved Information 3: ...
Thought 3: The event in Norfolk, Virginia, where Romney announced Ryan as his running mate, is a specific campaign event, and the fact that it took place on August 11, 2012, is supported by multiple news sources. The 2012 Republican National Convention in Tampa, Florida, occurred later in August. These details align with the known timeline of events leading up to the convention. Therefore, the news caption appears to be accurate regarding the date, location.
Action 3: Finish[TEXT SUPPORT]
reward = 0.8
=======================Iter 2: select node <match>=====================
Thought 1: I need to determine what the image depicts to see if it aligns with the news caption about Romney announcing Ryan as his running mate in front of the USS Wisconsin.
Action 1: VQA[What is shown in the image?]
Observation 1: The image shows a person standing on a stage or platform, addressing or speaking to a large crowd of people gathered in front of them. The stage is outdoors, with a blue sky and some clouds visible in the background. The crowd appears to be listening attentively, and there are some microphones and audio equipment set up on the stage.
Thought 2: I need to verify whether the background or setting includes the USS Wisconsin or a similar naval vessel to confirm if the location matches the description in the caption.
Action 2: VQA[Are there any naval ships, specifically the USS Wisconsin, visible in the background or nearby in the image?]
Observation 2: No, there are no naval ships, including the USS Wisconsin, visible in the background or nearby in the image. The background shows a clear sky and some distant objects that do not appear to be ships.
Thought 3: The image does not show the USS Wisconsin or any naval vessel, and there is no evidence to suggest the setting matches the location described in the caption. Therefore, the image content does not support the news text.
Action 3: Finish[MISMATCH]
reward = 0.8
=======================Final decision===============================
Finish[MISMATCH]
\end{lstlisting}
\end{tcolorbox}
\end{figure*}

\begin{figure*}[ht]
\begin{tcolorbox}[
    colback=white, 
    colframe=red!60!black, 
    fonttitle=\bfseries,
    title=Bad case: Over-reasoning leads to incorrect text-image association,
    enhanced,
    attach boxed title to top left={yshift=-2mm,xshift=4mm},
    boxed title style={
        colback=red!60!black, 
        colframe=red!60!black, 
    },
    coltitle=white 
]
\noindent
\begin{tabular}{@{}c@{}}
    \includegraphics[width=0.45\textwidth]{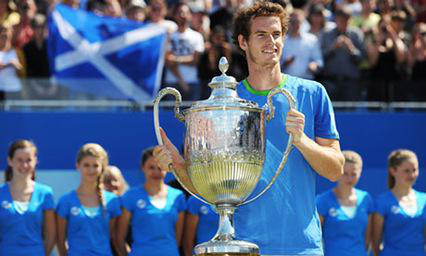}  
\end{tabular}
\hfill
\begin{tabular}{p{0.45\textwidth}}
    \textbf{News Text}: Andy Murray returns the ball to YenHsun Lu he won comfortably also in straight sets.\\
    \textbf{Binary label}: {Fake}\\
    \textbf{Multiple label}: {Mismatch}\\
\end{tabular}
\begin{lstlisting}[breaklines=true，basicstyle=\ttfamily\footnotesize, columns=fullflexible]
=======================Iter 0: select node <match>=====================
Thought 1: I need to determine what the image depicts to see if it aligns with the caption.
Action 1: VQA[What is shown in the image?]
Observation 1: The image shows a person holding a large trophy, dressed in sports attire, with a crowd and other athletes in the background, suggesting a tennis victory celebration. Given the context, it is likely a tennis-related event. The person in the image appears to be male with curly hair and is smiling, indicating a positive outcome related to a sporting achievement.
Thought 2: The actions of the characters need to be further confirmed.
Action 2: VQA[Is the person in the image a tennis player celebrating a victory?]
Observation 2: The person is celebrating a victory, likely after a tennis match.
Thought 3: Based on the context of celebrating with a trophy and sports attire, this image aligns with a tennis victory, supporting the news claim.
Action 3: Finish[MATCH]

reward=0.6
=======================Iter 1: select node <text>=====================
Thought 1: I need to verify the recent performance and match results of Andy Murray and YenHsun Lu to determine the accuracy of this statement. Andy Murray is a well-known professional tennis player, and the match details are specific, so I will verify if Murray played and won against YenHsun Lu in a recent match.
Action 1: Google[Andy Murray vs YenHsun Lu match result]
Observation 1: Retrieved Information 1: Andy Murray wins men's singles Olympics tennis gold, Aug 5, 2012 ... Having suffered a shock first-round defeat by 77th-ranked Yen-Hsun Lu ... After saving two break points in the opening game of the match, he...
    Retrieved Information 2:...
Thought 2: The search results indicate that Andy Murray has played against YenHsun Lu and won a match in straight sets. Multiple sources mention Murray's victory over Lu, including references to a match where Murray "crushes" Lu and descriptions of him winning easily. There is no evidence contradicting that Murray won the match comfortably in straight sets.
Action 2: Finish[TEXT SUPPORT]

reward=0.8
=======================Iter 2: select node <image>=====================
Thought 1: Detect["image.jpg"]
Action 1: Detect["image.jpg"]
Observation 1: The news image conforms to the objective facts because the image shows a person holding a trophy, presumably celebrating a victory or achievement. The person is smiling and appears to be in a celebratory mood. There are other individuals in the background, some of whom are also smiling and appear to be part of the celebration. 
Thought 2: According to the observation, the news image conforms to the objective facts.
Action 2: Finish[IMAGE SUPPORT]

reward=0.8
=======================Final decision===============================
Finish[ORIGINAL]
\end{lstlisting}
\end{tcolorbox}
\end{figure*}

\subsection{Prompts}\label{Appendix:prompt}
We present the prompts used in our method, including: the reasoning trajectory score prompt and the result confidence score prompt used during evaluation; the initialization prompt, which is employed to score nodes during the initialization of the search process; and the task prompt, guiding the reasoning process throughout tree expansion. We take MMfakebench as an example.

{
\small

[a1] Chen, Y.; Li, D.; Zhang, P.; Sui, J.; Lv, Q.; Tun, L.; and Shang, L. 2022. Cross-modal ambiguity learning for mul-timodal fake news detection. In WWW.

[a2] Liu, X.; Liu, Y.; Chen, J.; and Liu, X. 2022. PSCC-Net: Progressive spatio-channel correlation network for image manipulation detection and localization. IEEE TCSVT.

[a3] Wu, Y.; Zhan, P.; Zhang, Y.; Wang, L.; and Xu, Z. 2021. Multimodal fusion with co-attention networks for fake news detection. In ACL-IJCNLP.
}
\begin{figure*}
\begin{tcolorbox}[
colback=white,
colframe=color1,
title=Reasoning Trajectory Score Prompt,
]
\textbf{Task:} Analyze Misinformation Detection Trajectories.
The trajectories are labeled by environmental observations about the situation, thoughts that can reason about the current situation, and actions. Given a news item and a trajectory, evaluate its correctness and provide your reasoning and analysis in detail. Focus on the latest thought, action, and observation.

\begin{enumerate}[label=(\arabic*),leftmargin=*]
\item Incomplete trajectories can be correct if the thoughts and actions so far are correct, even if the answer is not found yet. 
\item Do not generate additional thoughts or actions beyond those provided.
\item At the last line of your analysis, conclude with "Thus the correctness score is {s}", where s is an integer from 1 to 10.
\end{enumerate}
\end{tcolorbox}
\end{figure*}

\begin{figure*}
\begin{tcolorbox}[
colback=white,
colframe=color2,
title=Result Confidence Score Prompt,
]
\textbf{Task:} Given a news, thoughts, observations, and a generated answer. If the answer can be drawn by thoughts or observation, then the result is relatively reliable; otherwise, it is unreliable. Give a brief analysis of the reliability of the answer. Then, at the last line conclude "Thus the reliability score is {s}", where s is an integer from 1 to 10.
\end{tcolorbox}
\end{figure*}

\begin{figure*}
\begin{tcolorbox}[
colback=white,
colframe=color3,
title=Initialization Prompt,
]
\textbf{Task:}
Given a news text and a news image, your task is to infer the probability that the news belongs to the following three different types of forgery based on your experience:
\begin{enumerate}[label=(\arabic*),leftmargin=*]
\item  \textit{Textual Veracity Distortion}: 
If the news text contains rich information, the news is more likely to be text veracity distortion. 

\item  \textit{Visual Veracity Distortion}: 
If the content in the news image contains counterfactual scenarios (e.g. violates the physical laws), the news is more likely to be visual veracity distortion.

\item  \textit{Cross-modal Consistency Distortion}: 
If the content in the news image is irrelevant to the image or does not support the text, the news is more likely to be cross-modal consistency distortion.

\end{enumerate}

Please avoid redundant analysis and directly return the judgment result in the following form: "Thus, the possibility of Textual Veracity Distortion, Visual Veracity Distortion, Cross-modal Consistency Distortion are $[p_1,p_2,p_3]$", where $p_1,p_2,p_3$ are floats from 0 to 1.
\end{tcolorbox}
\end{figure*}

\begin{figure*}
\begin{tcolorbox}[
colback=white,
colframe=color4,
title=Task Prompt,
]
\textbf{Task 1: Textual Veracity Detection}
Solve a textual veracity detecting task with interleaving \textit{Thought}, \textit{Action}, and \textit{Observation} steps. You need to verify the knowledge-based information therein, such as public figures, political events, and scientific common sense.

\begin{itemize}[leftmargin=*]
\item \textit{Thought}: Can reason about the current situation.
\item \textit{Actions}:
\end{itemize}

\begin{enumerate}[label=(\arabic*),leftmargin=*]
\item \texttt{Wikipedia[entity]}: Searches the exact entity on Wikipedia and returns the first paragraph if it exists. If not, it will return some similar entities to search.

\item \texttt{Google[entity]}: Searches information on Google and returns the snippet if it exists. Please give priority to \texttt{Wikipedia}; \texttt{Google} should be considered when \texttt{Wikipedia} fails.

\item \texttt{Finish[answer]}: Return the answer and finishes the task.

If there is any credible objective evidence refuting the news caption, please answer in the form: \texttt{Finish[TEXT REFUTE]}. 
If no such evidence is found, please answer in the form: \texttt{Finish[TEXT SUPPORT]}.
\end{enumerate}

\textbf{Task 2: Image Veracity Detection}
Solve an image veracity detecting task (determine whether the content in the news image contains counterfactual scenarios, e.g., violates physical laws) with interleaving \textit{Thought}, \textit{Action}, and \textit{Observation} steps.

\begin{itemize}[leftmargin=*]
\item \textit{Thought}: Can reason about the current situation.
\item \textit{Actions}:
\end{itemize}

\begin{enumerate}[label=(\arabic*),leftmargin=*]
\item \texttt{Detect[image]}: Return the forgery situation detected by a forgery detection model.

\item \texttt{Finish[answer]}: Return the answer and finishes the task.

If there is any credible objective fact refuting the news image, please answer in the form:
\texttt{Finish[IMAGE REFUTE]}. 
If no such fact is found, please answer in the form:
\texttt{Finish[IMAGE SUPPORT]}.
\end{enumerate}

\textbf{Task 3: Cross-modal Matching Detection}
Solve a cross-modal matching detection task (determine whether the content in the news image supports the text or not) with interleaving \textit{Thought}, \textit{Action}, and \textit{Observation} steps. Please make a direct judgment based on the image content and the text content. 

\begin{itemize}[leftmargin=*]
\item \textit{Thought}: Can reason about the current situation.
\item \textit{Actions}:
\end{itemize}

\begin{enumerate}[label=(\arabic*),leftmargin=*]
\item \texttt{VQA[question]}: Return the description of the image information concerned in the question.

\item \texttt{Entity[image]}: Return the entity of the image, including the identity of public figures. 

\item \texttt{Finish[answer]}: Return the answer and finishes the task.

If the news caption matches the content of the news image, please answer in the form:
 \texttt{Finish[MATCH]}. If no match is found, please answer in the form: \texttt{Finish[MISMATCH]}.
\end{enumerate}
\end{tcolorbox}
\end{figure*}


\end{document}